\documentclass[letterpaper]{article} 
\usepackage{aaai24}  
\usepackage{times}  
\usepackage{helvet}  
\usepackage{courier}  
\usepackage[hyphens]{url}  
\usepackage{graphicx} 
\usepackage{color,xcolor}
\usepackage{amsthm,amsmath,amssymb}
\usepackage{mathrsfs}
\usepackage{multirow}
\usepackage{fancyhdr}

\usepackage{natbib}  
\usepackage{caption} 
\usepackage{algorithm}
\usepackage{algorithmic}

\usepackage{newfloat}
\usepackage{listings}
\DeclareCaptionStyle{ruled}{labelfont=normalfont,labelsep=colon,strut=off} 
\floatstyle{ruled}
\newfloat{listing}{tb}{lst}{}
\floatname{listing}{Listing}
\nocopyright 

\title{Time-Series Contrastive Learning against False Negatives and Class Imbalance}
\author{
    Xiyuan Jin \textsuperscript{\rm 1},
    Jing Wang \textsuperscript{\rm 1,2}\thanks{Corresponding author\\This work has been submitted to the IEEE for possible publication. Copyright may be transferred without notice, after which this version may no longer be accessible.},
    Lei Liu \textsuperscript{\rm 1},
    Youfang Lin \textsuperscript{\rm 1,2}
}

\affiliations{
    \textsuperscript{\rm 1}School of Computer and Information Technology, Beijing Jiaotong University, Beijing, China\\
    \textsuperscript{\rm 2}Beijing Key Laboratory of Traffc Data Analysis and Mining, Beijing, China\\

    \{xiyuanjin, wj, lei\_liu, yflin\}@bjtu.edu.cn;
}

\usepackage{bibentry}

\begin{document}

\maketitle

\begin{abstract}
As an exemplary self-supervised approach for representation learning, time-series contrastive learning has exhibited remarkable advancements in contemporary research.
While recent contrastive learning strategies have focused on how to construct appropriate positives and negatives, in this study, we conduct theoretical analysis and find they have overlooked the fundamental issues: \emph{\textbf{false negatives}} and \emph{\textbf{class imbalance}} inherent in the InfoNCE loss-based framework. 
Therefore, we introduce a straightforward modification grounded in the SimCLR framework, universally adaptable to models engaged in the instance discrimination task.
By constructing instance graphs to facilitate interactive learning among instances, we emulate supervised contrastive learning via the multiple-instances discrimination task, mitigating the harmful impact of false negatives. 
Moreover, leveraging the graph structure and few-labeled data, we perform semi-supervised consistency classification and enhance the representative ability of minority classes.
We compared our method with the most popular time-series contrastive learning methods on four real-world time-series datasets and demonstrated our significant advantages in overall performance.
\end{abstract}

\section{Introduction}
Benefiting from advancements in computational power and the availability of large-scale datasets, time-series-based deep learning has seen remarkable progress recently and finds extensive applications in disease diagnosis \cite{tang2021self}, traffic analysis \cite{guo2021learning}, and financial prediction \cite{ding2015deep}. However, annotating time-series in real-world scenarios poses unique challenges compared to other data types, such as images and videos, making large-scale dataset collection difficult. Specifically, in physiological time-series annotation, it is even necessary to involve multiple experts working collaboratively to ensure precise and reliable annotations \cite{gogolou2018comparing}.

As an important stepping stone towards generic representation learning without supervision, contrastive learning paradigms based on instance discrimination \cite{wu2018unsupervised,ye2019unsupervised} demonstrated exceptional performance, even surpassing current supervised methodologies across numerous tasks \cite{he2020momentum,chen2020simple}. 
Its primary objective is to learn an encoder to discriminate positive pairs from negative pairs without any notion of semantic categories and end up with representations that capture apparent similarity among instances.

Inspired by such a strategy, time-series contrastive learning (TCL) methods have emerged and proposed various sample construction strategies tailored to time-series data with InfoNCE loss \cite{oord2018representation}, such as contrastive predictive coding \cite{oord2018representation,eldele2021time}, subseries discrimination \cite{franceschi2019unsupervised}, temporal neighborhood discrimination \cite{tonekaboni2021unsupervised}, .etc. 
However, our in-depth theoretical analysis points out that current TCL methods still struggle with two fundamental challenges:

\textbf{C1: How to alleviate the impact of false negatives under the instance discrimination task?}

In the context of the instance discrimination task, unsupervised contrastive learning (UCL) often assumes manually designed augmentations or other views/modalities as positives, with the remaining samples within a mini-batch serving as negatives regardless of their semantic content \cite{chen2020simple}. 
Undoubtedly, this simple and efficient approach has greatly facilitated the widespread adoption of contrastive learning \cite{chen2020big,li2021contrastive}. 
However, ensuring the genuineness and effectiveness of negative samples remains a significant challenge. 
Without supervision, negative sample pairs are highly likely to consist of semantically similar or identical samples, which we refer to as false negatives. 
The existence of false negatives will severely impede the convergence of feature representation learning \cite{huynh2022boosting, zheng2021weakly}. 

Recent works related to this question \cite{huynh2022boosting} explored identifying false negatives and executing false negative elimination and attraction. However, the specific harmful impact of false negatives still lacks effective theoretical support, which motivates us to explore the theoretical basis and practicable solutions. 

\textbf{C2: How to improve the representation learning for highly imbalanced time-series classification?}

Existing UCL methods typically operate agnostically to the distribution of training data. Nevertheless, time-series data, particularly physiological time-series collected in real-world environments, often exhibit imbalanced distributions \cite{perslev2019u}.
This arises from the fact that disease occurrences typically endure shorter durations than non-disease periods, or some rare diseases inherently prevail less frequently than common ailments \cite{nattel2002new,banerjee2009descriptive}.  Consequently, the availability of specific-class time-series data becomes inherently limited \cite{cao2013integrated}. 
Regrettably, the ability of most UCL methods would be declining for pivotal but infrequent minority classes \cite{zeng2023imgcl}. Thus, a generic representation learning method and further research are needed.

Drawing inspiration from recent developments in supervised contrastive learning (SCL), this paper first theoretically analyzes the lower bound and dilemma of the InfoNCE loss in UCL. Then, we propose \emph{a \underline{\textbf{S}}emi-supervised \underline{\textbf{I}}nstance-graph-based \underline{\textbf{P}}seudo-\underline{\textbf{L}}abel \underline{\textbf{D}}istribution \underline{\textbf{L}}earning framework} (SIP-LDL) to mitigate the impact of false negatives while significantly improving the classification performance of minority classes. 
Specifically, we put forward a multiple-instances discrimination task to implement an approximation from UCL to SCL and further alleviate the impact of false negatives. 
Moreover, we propose to leverage an instance graph convolution to strengthen inter-instance relationships to replace the common-used linear projection head. 
Based on the feature propagation on the graph, we infer that the minority class suffers from more noisy neighbor instances than the majority leading to the feature underfitting, which motivated us to devise a semi-supervised consistency classification loss with minimal labeled samples to improve the loss and performance of the minority. 
In summary, our contributions can be summarized as follows:
\begin{itemize}
\item \emph{New theoretical perspective}: we present a comprehensive theoretical analysis and derive the lower bound of contrastive loss to demonstrate the two primary challenges of UCL: \emph{\textbf{false negatives}} and \emph{\textbf{class imbalance}}, which implicitly deteriorates the quality of learned representations.
\item \emph{Simple and effective framework}: we introduce a novel SIP-LDL framework. Executing a multiple-instances discrimination task and propagating features on the instance graph, we alleviate the challenges of false negatives. Additionally, we employ a semi-supervised learning paradigm, utilizing a limited number of labeled samples to optimize the representative ability of minority classes.
\item \emph{Convincing results}: Extensive experiments demonstrate the effectiveness of our proposed model, surpassing the state-of-the-art on multiple physiological time-series datasets and achieving significant improvements, particularly for minority classes.
\end{itemize}

\section{Related work}

\subsection{Contrastive Learning with Time-series}
Various frameworks for contrastive learning have been proposed, which stem from the basic motivation of instance discrimination \cite{wu2018unsupervised}.
Without relying on any semantic supervisory information, instance-wise discrimination employs an encoder to strategically pull positives into closer alignment while concurrently enhancing the distinct separation among negatives.
Subsequently, this process enhances the discriminative representation of each instance \cite{chen2020simple}.
Expanding on this point, current time-series contrastive works study how to construct better positive pairs based on temporal invariance in time-series.
For instance, CPC \cite{oord2018representation} introduced a prediction-based context discrimination task to capture underlying shared information, followed by TSTCC \cite{eldele2021time}, which proposed cross-view discrimination with strong and weak augmentations. 
Additionally, TS2Vec \cite{yue2022ts2vec} introduced fine-grained multi-scale context discrimination. 
Nevertheless, these approaches still exhibit limitations in guaranteeing the dissimilarity of chosen negative samples and fail to effectively tackle the problem of minority class feature underfitting in imbalanced datasets.

\subsection{Contrastive Learning with False Negatives}
False negative samples hinder the model's effective convergence by forcing it to learn in the opposite direction of instances with similar or even identical semantics \cite{huynh2022boosting, zheng2021weakly}. 
Certain endeavors have sought to integrate analogous instances into model training to mitigate the repercussions stemming from false negatives. 
To illustrate, Zheng et al. \cite{zheng2021weakly} construct a nearest neighbor graph for each instance within a mini-batch, subsequently deploying a KNN-based multi-crop strategy to identify false negatives.  
More recently,  Huynh et al. \cite{huynh2022boosting} explored techniques for identifying false negatives without supervision and proceeded to explicitly eliminate detected instances or attract them as positives.
In contrast, our approach involves utilizing samples from both the majority and minority classes with similar semantic traits to enhance learning rather than completely eliminate them.

\subsection{Contrastive Learning with Class Imbalance}

Learning discriminative representations on imbalanced datasets is a challenging task. 
Various methods have emerged to improve performance in the realm of imbalanced classification.
Hybrid-PSC \cite{wang2021contrastive} introduces a novel hybrid network structure that combines contrastive learning with other techniques, leading to promising results.
BCL \cite{zhu2022balanced} presents a balanced contrastive loss that optimizes all classes for a balanced feature space, even in the presence of long-tail distributions.
ImGCL \cite{zeng2023imgcl} utilizes the node centrality-based PBS method to automatically balance representations learned from GCL, which is particularly useful for graph-structured datasets.
However, these methods either heavily rely on supervised information or are tailored to specific data types, such as graphs. 
Therefore our method presents a dedicated endeavor towards ameliorating class imbalance leveraging a cost-effective paradigm concurrently applicable across any data type. 

\section{Perspective}
In the time-series classification task, we aim to learn a complex function $\varphi$ mapping from an input space $\mathcal{X}$ to the target space $\mathcal{Y} = [C] = \{1, 2, . . . , C\}$. The function $\varphi$ is usually implemented as the composition of an encoder $f: \mathcal{X} \to\mathcal{H} \in \mathbb{R}^{h}$ and a linear classifier $W: \mathcal{H} \to Y$. 
Usually, contrastive loss is applied on the embeddings $Z$ after a projection $g: \mathcal{H} \to\mathcal{Z} \in \mathbb{R}^{h}$.
Since the final classification accuracy strongly depends on the quality of the representations $Z$, we aim to learn an effective encoder $f$ to improve representations against false negatives and class imbalance.

Next, we will give an in-depth analysis to point out the two major issues currently troubling UCL: \emph{\textbf{false negatives}} and \emph{\textbf{class imbalance}} from the differences in losses between SCL and UCL from a multi-class classification perspective.

\subsection{Contrastive learning as Multi-Class classification}
Given any two views, we can interpret contrastive learning as multi-class classification operating their representations $(z_i, z_j)$ with a label 1 if it is sampled from the joint distribution $(i, j) \sim P_{ij}$, e.g., augmentations or other same-class samples, and -1 if it comes from the product of marginals, $(i, k) \sim P_iP_k$, e.g., other samples in a mini-batch. In the presence of false negatives, some negative pairs $(i, k) \sim P_iP_k$ should be labeled as positive.
\subsubsection{Supervised and unsupervised contrastive loss.} 
For an instance $x_i$ of representation $z_i$ in a batch $B$, supervised contrastive loss $\mathcal{L}_{\text{sc}}$ and unsupervised contrastive loss $\mathcal{L}_{\text{uc}}$ have the following expressions:

\begin{align}
    &\mathcal{L}_{\text{sc}}(i) = -\frac{1}{|B_{y_i}|-1}\sum_{j\in B_{y_i} \setminus \{ i\} 
 }\log\frac{\exp(z_i \cdot z_j/\tau)}{\sum_{k \in B \setminus \{i\}}\exp( z_i \cdot z_k/\tau)} 
\label{equation:sc}
\end{align}

\begin{equation}
    \mathcal{L}_{\text{uc}}(i) = -\log\frac{\exp(z_i \cdot z_j / \tau)}{\sum_{k \in B \setminus \{i\}}\exp(z_i \cdot z_k / \tau)}
\label{equation:uc}
\end{equation}
where $B_y$ is a subset of $B$ that contains all samples of class $y$, $|\cdot|$ stands for the number of samples in the set, $\tau > 0$ is a scalar temperature hyperparameter. 
The UCL and SCL functions involve calculating expectations over either single or multiple positive pairs $(i, j) \sim P_{ij}$, along with independent negative pairs $(i, k) \sim P_iP_k$. Their mathematical formulation takes the shape of the $(|B|-1)$-way softmax cross-entropy loss. In SCL models, the ultimate objective revolves around classifying the nature of multiple same-class pairs $\{(z_i,z_j)|j\in \{B_{y_i} \setminus \{ i\} \}\}$ as positives. In contrast, the UCL model focuses on the identification of the individual augmented pair $(z_i,z_j)$ as positive.

It is worth mentioning that we do not distinguish between the negative sample size in SCL and UCL, although we usually select $2|B|-2$ of the augmented samples as negatives with \emph{NT-Xent} paradigm in UCL \cite{chen2020simple}, which does not exert any discernible influence on the ultimate analysis. Note that we will further omit $\tau$ and use $\langle z_i,z_j\rangle$ instead for calculating the similarity of embeddings $z_i$ and $z_j$ for simplicity.

We introduce the class-specific batch-wise loss as follows:
\begin{equation}
    \mathcal{L}_{\text{sc}}(Z;Y,B,y) = \sum_{i\in B_y}  \mathcal{L}_{\text{sc}}(i)
\end{equation}
\begin{equation}
    \mathcal{L}_{\text{uc}}(Z;Y,B,y) = \sum_{i\in B_y}  \mathcal{L}_{\text{uc}}(i)
\end{equation}

\subsection{Two Main Drawbacks Analysis} 
In fact, previous work \cite{zhu2022balanced} has demonstrated that existing SCL forms an undesired asymmetric geometry configuration for imbalanced datasets.
In the following, we will give an in-depth analysis on UCL loss lower bound and prove that it leads to false negatives confliction as well as class imbalance issues homologous to the SCL loss. 

\emph{\textbf{Theorem 1.} Assuming the normalization function is applied for feature embeddings, let $Z = (z_1,\dots , z_N ) \in \mathbb{Z}^N$ be an N point configuration with labels $Y = (y_1, \dots , y_N ) \in [C]^N$, where $ Z = \{z \in \mathbb{R}^h : ||Z|| = 1\}$. The class-specific batch-wise SCL loss $\mathcal{L}_{\text{sc}}$ is bounded by:}
\begin{equation}
\begin{split}
&\mathcal{L}_{\text{sc}}(Z;Y,B,y) \ge \sum_{i\in B_y} \log( \underbrace{|B_y \setminus \{i\}|}_{constant \; term} + \\  &\underbrace{|B_y^C|\exp(\frac{1}{|B_y^C|}\sum_{k\in B_y^C} \langle z_i,z_k \rangle - \frac{1}{|B_y|-1}\sum_{j\in B_y \setminus \{i\}} \langle z_i,z_j \rangle}_{confrontation \; term}) )
\end{split}
\label{equation:sc_bound}
\end{equation}
where we define $B^C_y$ as the complement set of $B_y$.

The above lower bound of SCL loss is derived by \cite{graf2021dissecting}, which consists of a \emph{constant term} and a \emph{confrontation term} redefined in our analysis.
It means that the contrastive loss for class $y$ depends on the size of $y$ in the constant term and the size of its complement in the confrontation term that represents the confrontation between the mean representations of negative and positive sample pairs.

Furthermore, we have demonstrated in the Supplementary Material the differences in such losses between classes that the lower bound of SCL loss in the majority is much more than the minority, which leads to more beneficial gradient returns and better representations. 

\begin{figure*}[!h]
\centering
\includegraphics[width=1\linewidth]{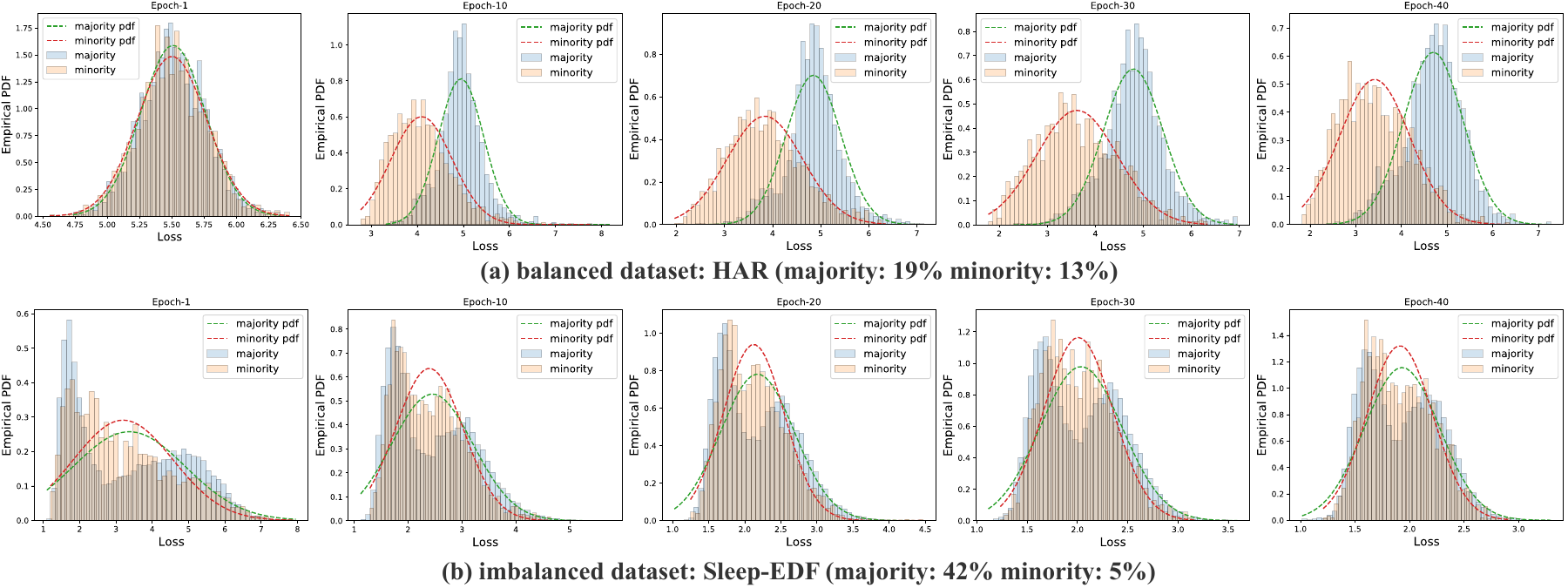}
\caption{
The loss distributions of the majority and minority on the balanced/imbalanced dataset as the training progresses.
}
\label{fig:imbablance_loss}
\end{figure*}

As shown in Figure \ref{fig:imbablance_loss}, we discern that a relatively balanced dataset allows effective differentiation between majority and minority classes after sufficient learning of several epochs, as indicated by the higher losses experienced by the majority classes. 
However, a severely imbalanced dataset reveals a more troubling revelation: the loss of the majority class closely approximates that of the minority class. 
We infer that the class imbalance is exacerbated during learning, which leads to a greater loss decline of the majority while impeding the feature learning of the minority. 

\emph{\textbf{Theorem 2.} Let Z, Y be defined as in Theorem 1, the class-specific batch-wise UCL $\mathcal{L}_{\text{uc}}$ loss is bounded by:}

\begin{equation}
\begin{split}
&\mathcal{L}_{\text{uc}}(Z;Y,B,y) \ge \sum_{i\in B_y} \log( \\
& \underbrace{|B_y \setminus \{i\}| \exp(\frac{1}{|B_y \setminus \{i\}|}\sum_{k\in |B_y \setminus \{i\}|} \langle z_i,z_k \rangle - \langle z_i,z_j \rangle)}_{confliction \; term}  \\
& + \underbrace{|B_y^C|\exp(\frac{1}{|B_y^C|}\sum_{k\in B_y^C} \langle z_i,z_k \rangle - \langle z_i,z_j \rangle)}_{confrontation \; term} )
\end{split}
\label{equation:uc_bound}
\end{equation}
the \emph{Proof.} has been provided in the Supplementary Material.

The differences between Equation (\ref{equation:sc_bound}) and Equation (\ref{equation:uc_bound}) are: 
1) the \emph{confliction term} replaces the \emph{constant term};
2) the mean of the positive sample representation $\frac{1}{|B_y|-1}\sum_{j\in B_y \setminus \{i\}} \langle z_i,z_j \rangle$ is replaced by $\langle z_i,z_j \rangle$.

Overall, the main drawbacks of UCL loss lie in: 
\begin{itemize}
\item The exponential term in the \emph{confliction term} indicates the learning confliction between false negative samples and positive pairs.
\item The minority classes suffer from deficiencies in losses and gradients, which further leads to the feature underfitting homologous to SCL.
\end{itemize}

\section{Method}
\begin{figure*}[!h]
\centering
\includegraphics[width=0.86\linewidth]{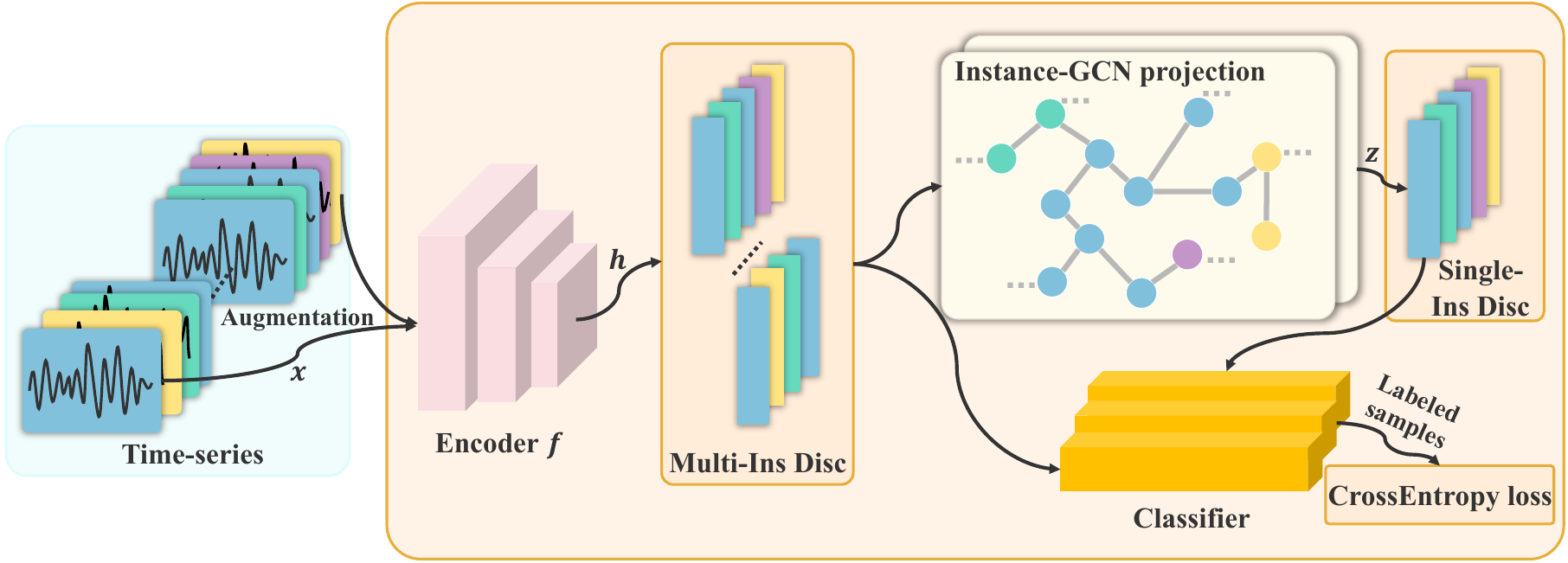}
\caption{
The overall architecture of the proposed SIP-LDL model.
}
\label{fig:main_model}
\end{figure*}
To address the two key issues in UCL mentioned above, we propose a simple but effective SIP-LDL framework as an extension of SimCLR \cite{chen2020simple}.

The overall architecture of the SIP-LDL is shown in Figure \ref{fig:main_model}. We propose to execute a multiple-instances discrimination task based on the single-instance discrimination task, which aims to implement an approximation from UCL to SCL and further alleviate the challenges of false negatives.
Then, we replace the projection head from MLP to GCN \cite{kipf2016semi} to enhance feature interaction learning.
On the one hand, instances of the majority can gain more substantial learning from same-class neighboring instances. 
On the other hand, losses of minority instances are enhanced through our semi-supervised consistency classification strategy as their features lack valid aggregation from scarce neighbor instances within the same class.

\subsection{Multiple-Instances Discrimination}
We get enlightenment from the loss form of SCL, which selects all instances of the same type in the entire mini-batch as positive samples to prevent the confliction of false negatives in the lower bound of SCL loss.

We aim to eliminate the gap between UCL and SCL as much as possible, which requires each instance to learn its corresponding pseudo-label distribution in the scenario of UCL. This means discovering other instances of the same type for each instance.
Specifically, based on the previous analysis, the contrastive loss based on InfoNCE is equivalent to the cross-entropy loss, and the embedding similarity between each instance and the rest $|B|-1$ samples can be used as the logical classification probability. 
Therefore, the goal of learning the distribution of instance pseudo labels is to discover and maximize the logical classification probability of potential samples of the same type, which is named multiple-instances discrimination in our study.

For a given time-series data, the pseudo-label distribution can be easily obtained by utilizing the similarity between samples.
To be specific, for a given input time-series sample $x_i$, we generate an augmentation set $\mathcal{T}$ through two separate augmentations (i.e., strong and weak augmentations) proposed by TSTCC \cite{eldele2021time}. 
Then, we feed its corresponding augmentations $x^a_i$ into a contrastive encoder $f_{enc}$ that maps samples to embeddings. We have $h_i=f_{enc}(x^a_i)$. 
Correspondingly, the logical classification probability $\alpha_{i,j}$ of sample $x_i$ for category $j$ (i.e., instance $x_j$) can be obtained from the following normalised similarity:
\begin{equation}
    \alpha_{i,j} = \frac{\exp(h_i \cdot h_j / \tau)}{\sum_{k \in B \setminus \{i\}}\exp(h_i \cdot h_k / \tau)}
\label{equation:similarity}
\end{equation}
Therefore, the multiple-instances discrimination loss of sample $x_i$ is as follows:
\begin{equation}
    \mathcal{L}_{\text{MID}}(i) =  - \frac{1}{|B|-1} \sum_{j \in B \setminus \{i\}} \log \frac{\exp(h_i \cdot h_j / \tau)}{\sum_{k \in B \setminus \{i\}}\exp(h_i \cdot h_k / \tau)}
\label{equation:mid}
\end{equation}
\subsubsection{Instance Graph Convolution}
While MLPs are commonly employed as projection heads for mapping representations into the space where contrastive loss is applied, it is important to note that discriminating multiple instances often yields suboptimal outcomes. 
This is primarily due to the absence of a collaborative learning process among instances, which results in numerous erroneous relationships within the similarity matrix.

Therefore, we make simple modifications to the original architecture. We first construct the instance graph based on the similarity matrix in Equation (\ref{equation:similarity}), then replace MLP with GCN, which can greatly enhance the learning of correlation between instances without increasing any parameter or changing the model architecture. 
To be specific, the embeddings $z_i$ of instance $x_i$ can be obtained after two-layer spatial GCN projections are as follows:

\begin{equation}
    z_i = \sum_{k \in B \setminus \{i\}}\alpha_{ij} (\sigma(\sum_{k \in B \setminus \{i\}}\alpha_{ij}h_{ij} W^{(1)} ))W^{(2)}
\end{equation}
where $\sigma(\cdot)$ denotes nonlinearity, $W^{(1)}$ and $ W^{(2)}$ are weight matrices. To ensure that the learning of instance nodes can retain their own characteristics after being propagated on the graph, we still perform single-instance discrimination after the GCN projection:
\begin{equation}
    \mathcal{L}_{\text{ID}}(i) = -\log\frac{\exp(z_i \cdot z_j / \tau)}{\sum_{k \in B \setminus \{i\}}\exp(z_i \cdot z_k / \tau)}
\end{equation}
where $z_i$ and $z_j$ are embeddings of positive pairs, and $z_i$ and $z_k$ are embeddings of negative pairs.

\subsection{Semi-supervised Consistency Classification}
As mentioned above, minority classes consistently struggle to obtain sufficient losses, regardless of whether in SCL or UCL, resulting in inadequate gradient assistance in learning representations.

Recently, SCL has shown promising performance in tackling imbalanced data through predefined class prototypes \cite{wang2021contrastive, zhu2022balanced}, to further enable each class to have an approximate contribution for optimizing. 
Nevertheless, acquiring complete data labeling does not align with the prerequisites of those employing UCL methodologies. 
Furthermore, enhancing loss outcomes for minority groups within the context of entirely UCL empirically proves to be a formidable challenge \cite{sun2023learning}. 
Therefore, we are driven to pursue a compromise approach: semi-supervised learning.

\begin{table}[!t]
\small
\begin{center}
    \begin{tabular}{c c c c c c }
    \hline
    Dataset & Train & Test & Length  & Class & $r_{im}$ \\
    \hline
    HAR & 7352 & 2947 & 128 & 6 & 1.44\\
    \hline
    Sleep-EDF & 25612 & 8910 & 3000 & 5 & 7.86 \\
    PhysioNet 2017 & 18256 & 5824 & 2500 & 4 &24.97\\
    TUSZ & 1925 & 521 & 2400 & 4 & 40.34    \\
    \hline
    \end{tabular}
\end{center}
\caption{Description of datasets used in our experiments.}
\label{table:datasets}
\end{table}

\begin{table*}
    \centering
    \setlength\tabcolsep{2.4pt}
    \begin{tabular}{l|cc|cc|cc|cc}
        \hline
        \multirow{2}{*}{Model Name}&\multicolumn{2}{c|}{HAR} &\multicolumn{2}{c|}{Sleep-EDF}  &\multicolumn{2}{c|}{PhysioNet 2017}&\multicolumn{2}{c}{TUSZ} \\
        \cline{2-9}
        & Accuracy &F1-score & Accuracy &F1-score & Accuracy &F1-score & Accuracy &F1-score \\
        \hline
        Supervised  & 92.22±0.63 & 92.22±0.67 & 83.29±0.62 &74.26±0.43 &54.00±3.22 &38.16±1.11 &\underline{63.42±0.64} &41.67±2.11  \\
        SCL         & 86.30±2.65 & 86.26±2.68 &\underline{84.23±0.38} &\textbf{75.01±0.30} & 56.42±0.11 & 21.96±0.63 & 58.81±0.93 & 26.62±4.26\\
        \hline
        SimCLR     &78.67±1.14 &78.01±0.94 &64.63±1.41 &54.21±1.15  &56.29±0.08 & 21.75±0.88&56.89±0.09 &18.29±0.32\\
        T-loss      &79.51±3.49 &78.78±3.83 &80.53±0.78 &70.15±0.32  &54.27±0.84 & 37.15±0.51 &62.03±0.53 &38.07±0.96\\
        CPC         &68.56±3.04 &68.32±3.22 &62.68±2.62 &47.71±3.10  &54.46±1.27 & 38.83±0.52 &61.77±0.43 &36.41±0.71\\
        TSTCC       &88.31±1.00 &88.31±1.02 &81.41±0.55 &70.53±0.29  &\underline{56.52±1.02}& \underline{40.48±0.50} &60.73±1.03 &35.36±3.98\\
        TS2Vec      &\underline{92.98±0.49} &\underline{93.01±0.50} &80.46±0.41 &69.04±0.55  &56.27±0.24 & 24.39±1.11 &62.50±0.77 &\underline{42.26±1.61}\\
        \hline
        MLP+$\mathcal{L}_{\text{ID}}$ & 79.95±1.73 & 81.28±1.70 & 65.28±1.57 & 57.49±2.97 &56.17±0.09 & 22.22±0.89& 56.78±0.19 & 18.11±0.04\\
        MLP+$\mathcal{L}_{\text{MID}}$ &77.33±3.29 & 76.63±3.71 & 72.58±0.87 & 62.72±1.37  &56.61±0.21 & 23.40±1.53 & 58.62±1.80 & 25.17±6.23 \\
        GCN+$\mathcal{L}_{\text{ID}}$ & 59.95±2.01 & 56.74±2.02 & 41.08±2.28 & 24.57±1.69 &56.01±0.47 & 20.71±0.49& 57.08±0.09 & 18.49±0.40\\
        GCN+$\mathcal{L}_{\text{MID}}$ &58.35±2.34 &53.96±2.60& 43.16±1.14 & 28.13±1.43 &56.28±0.03 & 20.02±1.44 &58.62±1.29 & 25.65±4.92\\
        MLP+$\mathcal{L}_{\text{MID}}$+$\mathcal{L}_{\text{ID}}$ &80.48±2.09 & 79.51±2.23&75.86±1.63 &64.80±1.49 &56.74±0.20 &25.50±0.83 & 59.58±1.42 & 29.92±5.63\\
        GCN+$\mathcal{L}_{\text{MID}}$+$\mathcal{L}_{\text{ID}}$ & 83.35±1.07 & 82.59±1.08 & 78.51±0.94 & 66.79±0.57 &58.23±0.29 & 28.64±0.63&  61.15±1.65 & 35.87±4.37\\
        \textbf{SIP-LDL}
                 &\textbf{93.20±0.66} &\textbf{93.27±0.64} &\textbf{84.32±0.35} & \underline{74.46±0.41} &\textbf{58.71±0.54} & \textbf{41.32±1.24} &\textbf{63.72±0.99} &\textbf{48.20±3.23}\\
        \hline
    \end{tabular}
\small\caption{Comparison between our proposed model against baselines using linear classifier evaluation experiment. The best results are marked in bold, the second-place results are underlined. }
\label{table:total}
\end{table*}

With a limited number of balanced labeled data, we train a classifier $f_C$ with representations before and after graph convolution constraints as follows:
\begin{equation}
    \mathcal{L}_{C}^{h}= - \sum_{l\in \mathcal{Y_L}} y_l log( f_C(h_l))
\end{equation}
\begin{equation}
    \mathcal{L}_{C}^{z}= - \sum_{l\in \mathcal{Y_L}} y_l log( f_C(z_l))
\end{equation}
where $\mathcal{Y_L}$ is the set of labeled instances, $\mathcal{L}_{CC} = \mathcal{L}_{C}^{h} + \mathcal{L}_{C}^{z}$ denotes the consistency classification loss.

The motivation behind our adoption of such a loss function stems from the realization that the minority class, following feature propagation on the graph, is inevitably susceptible to the influence of a multitude of noisy neighbor instances (i.e., majority classes), which resulted in suboptimal representations for the minority class. 
Therefore, the classification loss incurred on a uniformly sampled dataset inherently accentuates the loss experienced by the minority class, consequently engendering a more pronounced and effective gradient gain.

The final contrastive loss is the combination of the two multiple/single instance discrimination losses and the consistency classification loss as follows:
\begin{equation}
    \mathcal{L} =  \lambda_1 \cdot (\mathcal{L}_{\text{MID}} +  \mathcal{L}_{\text{ID}}) + \lambda_2 \cdot \mathcal{L}_{CC}
\end{equation}
where $\lambda_1$ and $\lambda_2$ are fixed scalar hyperparameters denoting the relative weight of each loss.


\section{Experimental Setup}

\subsection{Datasets}
To evaluate our model, we adopted 1 class-balanced dataset for human activity recognition and 3 class-imbalanced datasets for sleep stage classification, cardiac arrhythmia classification, and epileptic seizure classification, respectively. Detailed statistical information is summarized in Table \ref{table:datasets}. $r_{im}$ denotes the ratio of class-imbalance.

\subsubsection{Human Activity Recognition (HAR)}
We use UCI HAR dataset \cite{anguita2013public}, which contains sensor readings for 30 subjects performing 6 activities (i.e. walking, walking upstairs, downstairs, standing, sitting, and lying down). 

\subsubsection{Sleep Stage Classification (Sleep-EDF)}
We focus on Sleep Stage Classification, where we aim to categorize the input EEG signal into five classes: Wake (W), Non-rapid eye movement stages (N1, N2, N3), and Rapid Eye Movement (REM) stage \cite{goldberger2000physiobank}.  We used a single EEG channel (Fpz-Cz) with a sampling rate of 100 Hz.

\subsubsection{Cardiac Arrhythmia Classification (PhysioNet 2017)}
PhysioNet 2017 \cite{clifford2017af} consists of 8,528 single-lead ECG recordings alongside 4 different classes: Normal, Atrial Fibrillation, Other Rhythm, and Noisy.

\subsubsection{Epilepsy Seizure Classification (TUSZ)}
The TUSZ dataset \cite{obeid2016temple} comprises a total of 5,612 EEGs, with 3,050 annotated seizures, including 19 EEG channels and covering four distinct seizure types: combined focal (CF), generalized non-specific (GN), absence (AB), and combined tonic (CT) seizures \cite{tang2021self}.

\subsection{Implementation Details}
Experiments were repeated 5 times with 5 different seeds, and we reported the mean and standard deviation. 
The pretraining and downstream tasks were done for 40 epochs as we noticed that the performance did not improve with further training. 
We applied a batch size of 128. 
We used the Adam optimizer with a learning rate of 3e-4 and weight decay of 3e-4. 
We set $\lambda_1 = \lambda_2 = 1$.
Lastly, we built our model using PyTorch 1.9 and trained it on an NVIDIA RTX A4000 GPU.
We leverage 10\% balanced labeled data to train semi-supervised consistency classification.

\section{Results}

\subsection{Comparison with Baseline Approaches}
We compare our proposed approach against the following baselines. 
(1) \textbf{Supervised}: supervised training of both encoder and classifier model; 
(2) \textbf{SCL}: supervised contrastive learning applied on SimCLR;
(3) \textbf{SimCLR};
(4) \textbf{T-loss};
(5) \textbf{CPC};
(6) \textbf{TSTCC};
(7) \textbf{TS2Vec}.
It is worth noting that we use strong and weak augmentations to adapt SimCLR to our application, as it was originally designed for images.

To evaluate the performance of our SIP-LDL, we follow the standard linear benchmarking evaluation scheme, which trains a linear classifier on top of a frozen pre-trained encoder. 
Table \ref{table:total} shows the linear evaluation results of our approach against the baseline methods. 
Overall, our proposed SIP-LDL outperforms all five state-of-the-art methods on both class-balanced and imbalanced datasets while achieving comparable performance to the supervised approaches on the Sleep-EDF dataset. 
This demonstrates the powerful representation learning capability of our model.

\subsection{Ablation Study}
To assess the individual contributions of each module in our scheme, we designed several variant models involving modifications to specific modules while keeping the rest of the architecture unchanged. 
In our study, we started by training a basic encoder+MLP model for the single-instance discrimination task referred to as MLP+$\mathcal{L}_{\text{ID}}$. 
Building on this, we replaced single-instance discrimination with multiple-instances discrimination, called MLP+$\mathcal{L}_{\text{MID}}$. 
Then, we replaced the MLP projection head with GCN, resulting in GCN+$\mathcal{L}_{\text{ID}}$ and GCN+$\mathcal{L}_{\text{MID}}$.
Additionally, we simultaneously conducted multiple-instances discrimination for encoder representations and single-instance discrimination for projections, represented as MLP+$\mathcal{L}_{\text{MID}}$+$\mathcal{L}_{\text{ID}}$ and GCN+$\mathcal{L}_{\text{MID}}$+$\mathcal{L}_{\text{ID}}$ respectively. 
Finally, we trained the SIP-LDL model with $\mathcal{L}_{CC}$ by incorporating a small amount of labeled data for consistency classification.

The results in Table \ref{table:total} show that merely using multiple-instances discrimination did not consistently improve performance, likely due to limited interactions among instances. 
Introducing graph structures alone did not lead to significant improvements either. 
However, combining the instance graph and $\mathcal{L}_{\text{MID}}$+$\mathcal{L}_{\text{ID}}$  in our approach enhanced the model's performance.
Remarkably, by applying consistency classification $\mathcal{L}_{CC}$ with only 10\% of labeled data, we achieved substantial overall performance improvements.

\subsection{Class Analysis}
\begin{figure}[!h]
  \centering
  \includegraphics[width=\linewidth]{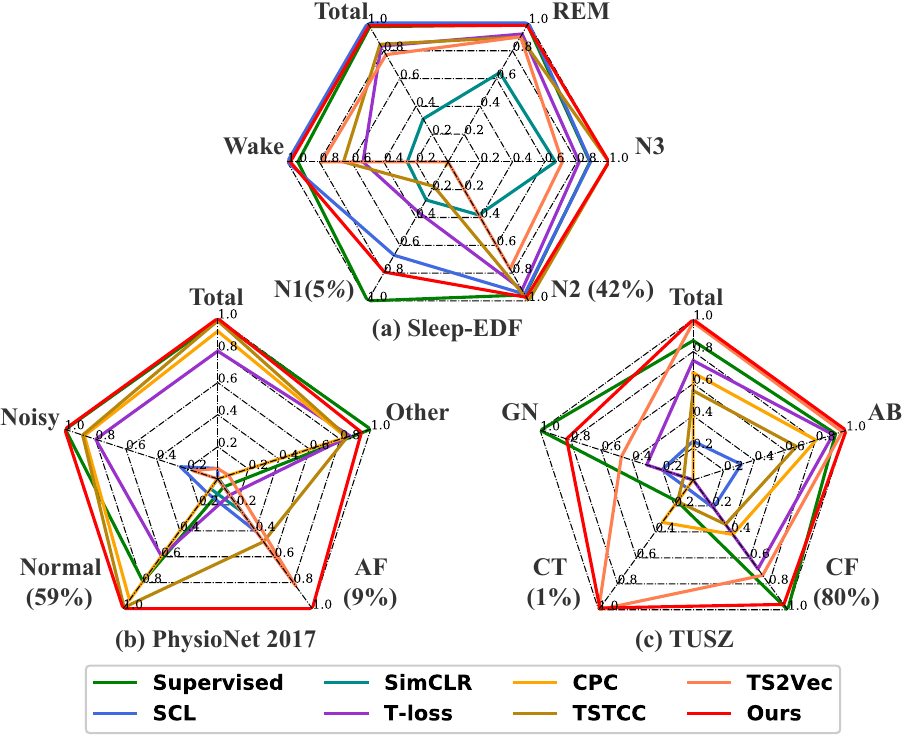}
  \caption{
    Normalized accuracy performance patterns of different sleep stages, cardiac arrhythmia types, and seizure types. The total and each class's F1 scores were evaluated. The result of the method with the highest accuracy was recorded as 1. (a) Sleep-EDF dataset. (b) PhysioNet 2017 dataset, (c) TUSZ dataset.}
  \label{fig:down_stream}
\end{figure}

\begin{figure}[!h]
  \centering
  \includegraphics[width=\linewidth]{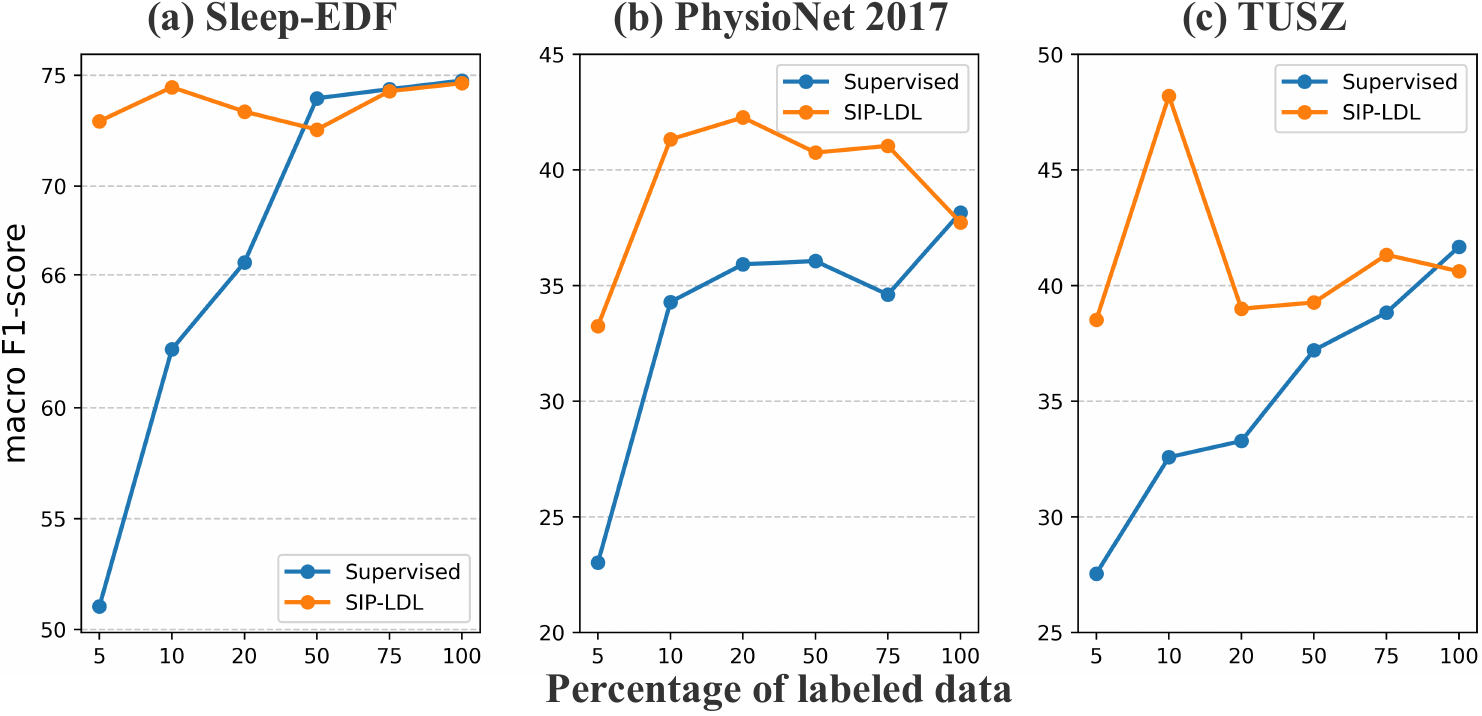}
  \caption{
    Comparison between supervised training vs. SIP-LDL for different few-labeled data scenarios in terms of MF1.}
  \label{fig:semi}
\end{figure}
To assess the quality of the generated embeddings, we conducted a comparative analysis of SIP-LDL against supervised methods and five contrastive learning baselines on downstream classification tasks, with a specific focus on three class imbalance datasets. In Figure \ref{fig:down_stream}, we present a comprehensive comparison of overall classification F1 scores and F1 scores for each class, along with indicating the proportions of samples with the maximum and minimum amounts within each dataset. 

Almost all the baseline methods perform poorly in the minority classes, while our SIP-LDL outperformed across all metrics, even surpassing supervised approaches in certain minority classes (e.g., AF in PhysioNet 2017 and CT in TUSZ). 
Compared to the classical SimCLR method based on the single-instance discrimination task, our proposed SIP-LDL strategy achieved substantial improvements in overall classification performance at the cost of limited labeled annotations.

\subsection{Semi-supervised Training}

We examined the effectiveness of SIP-LDL in a semi-supervised context. We trained the model with different percentages of labeled data: 5\%, 10\%, 20\%, 50\%, 75\%, and 100\%. 
SIP-LDL employs semi-supervised consistency classification, depicted by the orange curves in Figure \ref{fig:semi}. 
We found that supervised training struggles with limited labeled data. 
In contrast, our SIP-LDL method attains optimal results by employing a mere 10\% of labeled data while concurrently rivaling the supervised method with 100\% labeled data across three imbalanced datasets. 
While the supervised approach exhibit progressively improved performance with an increasing pool of available labeled instances, it is essential to underscore a notable divergence. 
Namely, when confronted with an escalation in labeled data, our SIP-LDL paradoxically experiences a performance decline.  
This phenomenon stems from the adverse effect of an enlarged label proportion on the integrity of our semi-supervised consistency classification framework. 
Consequently, the equilibrium among diverse sample classes is disrupted, leading to an encumbered learning process for the encoder and classifier, eventually succumbing to an imbalanced state.

\section{Conclusion}
We introduce a straightforward yet effective framework, SIP-LDL, to address the challenges of false negatives and class imbalance inherent in the single-instance discrimination task. 
Our proposed SIP-LDL employs a graph-based projection approach, imposing constraints on learning pseudo-label distributions for each instance, thereby approximating supervised contrastive learning. 
Simultaneously, we present a semi-supervised consistency classification loss, utilizing a mere 10\% of annotated data, leading to substantial enhancements in minority classes. Remarkably, our framework can be seamlessly integrated into existing contrastive learning models for single-instance discrimination tasks without introducing additional model parameters.
\bibliography{aaai24}

\end{document}


\section{A. Illustration of Decline about TCL methods on Imbalanced Dataset}

\begin{figure}[!h]
  \centering
  \includegraphics[width=\linewidth]{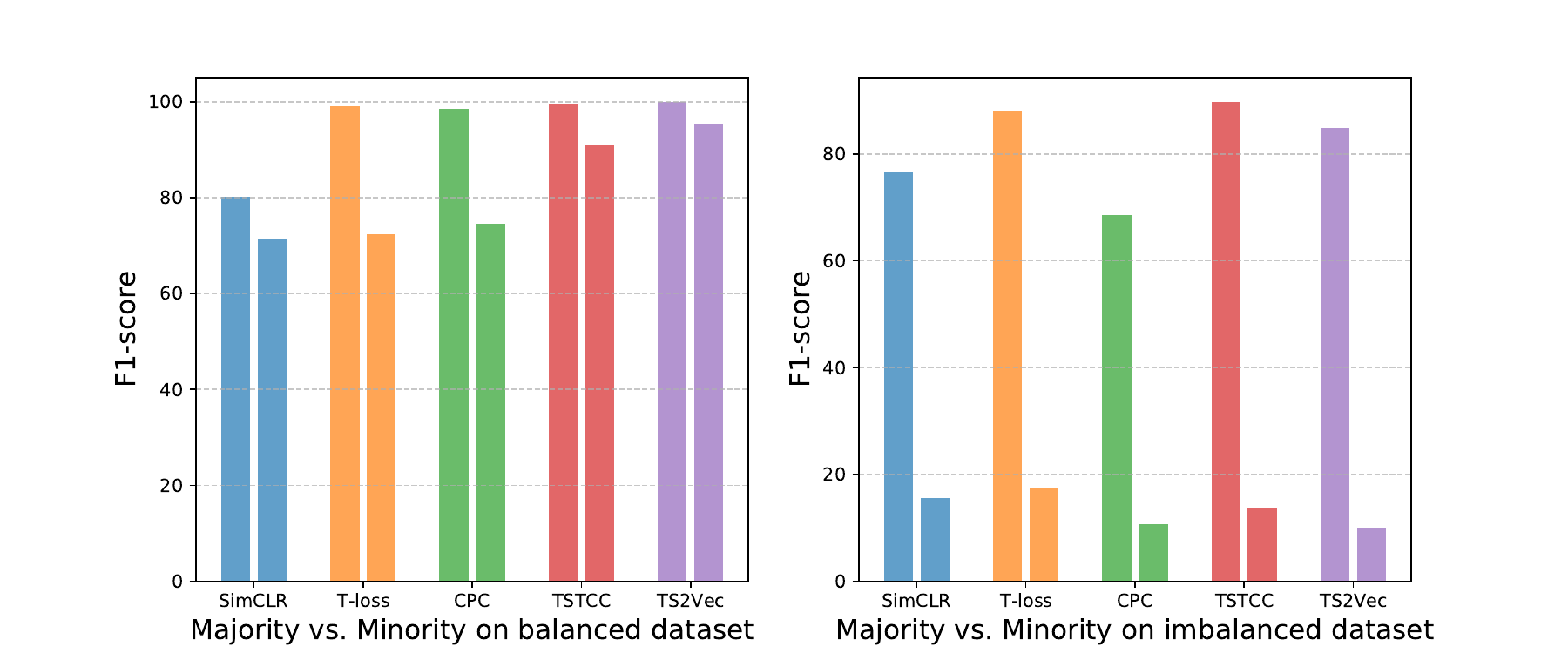}
  \caption{
    Comparison of the prevailing time-series contrastive learning baseline models on the balanced dataset (HAR) and imbalanced dataset (Sleep-EDF). Each method separately calculates F1-score for majority (left bar) and minority (right bar) classes.}
  \label{fig:deline}
\end{figure}

Shown as Figure \ref{fig:deline}, we compared the performance of current popular time-series contrastive learning (TCL) methods on balanced and imbalanced datasets to illustrate the current dilemma. Specifically, we separately counted the F1-score of the majority and minority classes for each method. On the balanced dataset, there is almost no significant difference in their performance across different classes. However, on imbalanced datasets, all methods exhibit a decline in performance in minority classes compared to majority classes.

\section{B. Proof of Theorem 1}
\begin{figure*}[!h]
\centering
\includegraphics[width=1\linewidth]{AnonymousSubmission/LaTeX/imbalance_loss.pdf}
\caption{
The loss distributions of the majority and minority on the balanced/imbalanced dataset as the training progresses.
}
\label{fig:imbablance_loss}
\end{figure*}
In the following, we provide proof for all technical details of Theorem 1.

\emph{\textbf{Theorem 1.} Assuming the normalization function is applied for feature embeddings, let $Z = (z_1,\dots , z_N ) \in \mathbb{Z}^N$ be an N point configuration with labels $Y = (y_1, \dots , y_N ) \in [C]^N$, where $ Z = \{z \in \mathbb{R}^h : ||Z|| = 1\}$. The class-specific batch-wise SCL loss $\mathcal{L}_{\text{sc}}$ is bounded by:}
\begin{equation}
\begin{split}
&\mathcal{L}_{\text{sc}}(Z;Y,B,y) \ge \sum_{i\in B_y} \log( \underbrace{|B_y \setminus \{i\}|}_{constant \; term} + \\  &\underbrace{|B_y^C|\exp(\frac{1}{|B_y^C|}\sum_{k\in B_y^C} \langle z_i,z_k \rangle - \frac{1}{|B_y|-1}\sum_{j\in B_y \setminus \{i\}} \langle z_i,z_j \rangle}_{confrontation \; term}) )
\end{split}
\label{equation:sc_bound}
\end{equation}
Such a lower bound for supervised contrastive learning (SCL) loss follows from an application of Jensen’s inequality. 
In particular, we first need to bring the class-specific batch-wise loss in a form amenable to Jensen’s inequality. Since $|B_y|>1$, we have that:
\begin{equation}
\begin{split}
\mathcal{L}_{\text{sc}}(Z;Y,B,y) &=  \sum_{i\in B_y} -\frac{1}{|B_{y_i}|-1} \\
& \sum_{j \in B_{y_i} \setminus \{i\} } 
\log (\frac{\exp(\langle z_i,z_j\rangle)}{\sum_{k \in B \setminus \{i\}} \exp(\langle z_i,z_k\rangle)) }) \\
& = \sum_{i\in B_y} \log( \frac{\sum_{k\in B \setminus \{i\}}\exp(\langle z_i,z_k\rangle)}{\prod_{j \in B_{y_i} \setminus \{i\}}\exp(\langle z_i,z_j\rangle)^{\frac{1}{|B_{y_i}|-1}} } ) \\
&= \sum_{i\in B_y} \log( \frac{\sum_{k\in B \setminus \{i\}}\exp(\langle z_i,z_k\rangle)}{\exp( \frac{1}{|B_{y_i}|-1} \sum_{j \in B_{y_i} \setminus \{i\}} \langle z_i,z_j\rangle) } )
\end{split}
\label{equation:lsc}
\end{equation}

We will focus on the sum in the numerator: For every $i \in B_y$, we write:

\begin{equation}
\begin{split}
\sum_{k\in B \setminus \{i\}}\exp(\langle z_i,z_k\rangle) =& \sum_{k\in B_y \setminus \{i\}}\exp(\langle z_i,z_k\rangle) \\
&+ \sum_{k\in B_y^C}\exp(\langle z_i,z_k\rangle)
\end{split}
\end{equation}

As the exponential function is convex, we can leverage Jensen’s inequality on both sums, resulting in:
\begin{equation}
\begin{split}
& \sum_{k\in B_y \setminus \{i\}}\exp(\langle z_i,z_k\rangle) \ge^{Q1} \\
&|B_y \setminus \{i\}| \exp( \frac{1}{|B_y \setminus \{i\}|} \sum_{k\in B_y \setminus \{i\}}  \langle z_i,z_k\rangle)
\end{split}
\end{equation}

\begin{equation}
\sum_{k\in B_y^C}\exp(\langle z_i,z_k\rangle) \ge^{Q2}  |B_y^C| \exp ( \frac{1}{|B_y^C|} \sum_{k\in B_y^C} \langle z_i,z_k\rangle )
\end{equation}

Herein, equality is attained if and only if:
\begin{itemize}
\item (Q1) There is a $C_i(B, y)$ such that $\forall j \in B_y \setminus \{i\} $ all inner products $\langle z_i, z_j \rangle = C_i(B, y)$ are equal. 
\item (Q2) There is a $D_i(B, y)$ such that $\forall j \in B_y \setminus \{i\}$ all inner products $\langle z_i, z_j \rangle = D_i(B, y)$ are equal.
\end{itemize}
Thus, using $\exp(a)/ \exp(b) = \exp(a-b)$, we bound the argument of the log in Equation (\ref{equation:lsc}) by:
\begin{equation}
\begin{split}
    &\mathcal{L}_{\text{sc}}(Z;Y,B,y) = \sum_{i\in B_y} \log( \frac{\sum_{k\in B \setminus \{i\}}\exp(\langle z_i,z_k\rangle)}{\exp( \frac{1}{|B_{y_i}|-1} \sum_{j \in B_{y_i} \setminus \{i\}} \langle z_i,z_j\rangle) } )\\
    & = \sum_{i\in B_y} \log( \frac{ \sum_{k\in B_y \setminus \{i\}}\exp(\langle z_i,z_k\rangle) + \sum_{k\in B_y^C}\exp(\langle z_i,z_k\rangle)}{\exp( \frac{1}{|B_{y_i}|-1} \sum_{j \in B_{y_i} \setminus \{i\}} \langle z_i,z_j\rangle) } ) \\
    &\ge \sum_{i\in B_y} \log( \underbrace{|B_y \setminus \{i\}|}_{constant \; term} + \\  &\underbrace{|B_y^C|\exp(\frac{1}{|B_y^C|}\sum_{k\in B_y^C} \langle z_i,z_k \rangle - \frac{1}{|B_y|-1}\sum_{j\in B_y \setminus \{i\}} \langle z_i,z_j \rangle}_{confrontation \; term}) )
\end{split}
\end{equation}
Note that equality is attained if and only if the above conditions hold for every $i \in B_y$. Also, note that the respective constants, $C_i(B, y)$ and $D_i(B, y)$, depend indeed on the batch $B$ and the label $y$.

\section{C. Imbalance in SCL loss}

For a $C$-way time-series classification problem, let $\{x_i, y_i\}^N_{i=1}$ be an imbalanced training set. The total number of the training set over $C$ classes is $N = \sum^C_{c=1} N_c$, where $N_c$ denotes the number of samples in class $c$. Let $\pi$ be the vector of label frequencies, where $\pi_c = \frac{N_c}{N}$ denotes the label frequency of class $c$. 
Without loss of generality, we assume that the classes are sorted by $\pi_c$ in descending order (i.e., if the class index $i < j$, then $N_i \geq N_j$, and $N_1 \gg N_C$). We denote by $r_{im} = \frac{N_1}{N_C}$ the imbalance ratio of the dataset. Therefore, the lower bound of SCL losses of the majority and minority classes $LB_{\text{sc}}(Z;Y,B,y_1)$ and $LB_{\text{sc}}(Z;Y,B,y_C)$ can be described in the following form
\begin{equation}
\begin{split}
    LB_{\text{sc}}(Z;Y,B,y_1) &= \sum_{i\in B_y} \log(r_{im}N_C + N_C e_1) \\
    &= \sum_{i\in B_y} \log((r_{im}+e_1)N_C )\\
\end{split}
\end{equation}
\begin{equation}
\begin{split}
    LB_{\text{sc}}(Z;Y,B,y_C) &= \sum_{i\in B_y} \log( N_C + r_{im}N_Ce_C) \\
    &= \sum_{i\in B_y} \log( (1+r_{im}e_C)N_C)
\end{split}
\end{equation}
where $e_1,e_C$ denotes the exponential term in the Equation (\ref{equation:sc_bound}). 
Let's consider the situation where the representations of minority and majority classes are not fully learned in the early stages of training, i.e., $e_1 \approx e_C = e \leq 1$, which means the impact of the repulsion term and attraction term is almost the same. 
It is worth mentioning that the more complete one learns, the smaller the exponential term.
This causes the difference between majority and minority classes within the logarithmic term to be as follows:
\begin{equation}
    (r_{im}+e)-(1+r_{im}e)=(r_{im}-1)(1-e) \geq 0
\end{equation}
Therefore we can infer that:
\begin{equation}
    LB_{\text{sc}}(Z;Y,B,y_1) \geq LB_{\text{sc}}(Z;Y,B,y_C)
\end{equation}
The above equation indicates that in the early stages of training, the majority class has more losses and gradients than the minority class.
It can be envisaged that in SCL scenarios, the representations of the majority of classes will progressively improve, while the representations of minority classes will gradually deteriorate.

\section{D. Detailed Explanation of the Imbalance Phenomenon}
In order to illustrate the limitations of class imbalance in SCL loss, we provide a more detailed supplement to the main text. We have plotted the contrastive loss distribution of majority and minority classes on relatively balanced and imbalanced datasets with training rounds progressing, as shown in Figure \ref{fig:imbablance_loss}. 
In addition, we have provided the proportion of majority and minority class samples in each dataset.

Specifically, we found that in the early stages of training (i.e., epoch 1), there was no significant difference in the distribution of majority and minority classes, whether on relatively balanced or imbalanced datasets. 
However, as the training rounds processes (i.e., epoch 10-40), the performance of the difference in the loss functions between the two classes on a balanced dataset means that when the representation of the majority and minority classes has almost the same expressive ability, their lower bound of class-wise contrastive loss is influenced by the specific class sample size. 
That is, the lower bound of the contrastive loss function of the majority class is higher than that of the minority class. 
Therefore, we observed that the class-wise contrastive loss of the majority class samples is ultimately higher than that of the minority class samples.

However, the performance varies greatly on the imbalanced dataset. We found that there is no significant loss difference between majority and minority classes as training progresses. Importantly, it should be noted that this does not mean that the representations of the majority and minority classes have the same expressive power. 
On the contrary, we believe that the phenomenon of almost identical loss distribution is precisely due to the fact that the representation of majority classes is better learned than that of minority classes, resulting in faster loss optimization on majority classes, which can be confirmed from the final downstream classification task on Table 1-4.

Overall, we believe that the difference in loss distribution among the different datasets mentioned above is the main reason for the poor performance of existing time-series contrastive methods on imbalanced datasets. 
Furthermore, this almost identical loss distribution also prevents us from using the commonly used class imbalance resolution strategy that relies on loss differences. 
Furthermore, the almost identical loss distribution also prevents us from learning and consulting from the commonly used class imbalance resolution strategies that rely on loss differences.
Further, under the setting of unsupervised contrastive learning, this dilemma is even more difficult to solve. 
Therefore, our SIP-LDL framework adopts a semi-supervised experimental setup to provide inspiration for further research in the future.

\section{E. Proof of Theorem 2}
In the following, we provide proof for all technical details of Theorem 2.

\emph{\textbf{Theorem 2.} Let Z, Y be defined as in Theorem 1, the class-specific batch-wise UCL $\mathcal{L}_{\text{uc}}$ loss is bounded by:}

\begin{equation}
\begin{split}
&\mathcal{L}_{\text{uc}}(Z;Y,B,y) \ge \sum_{i\in B_y} \log( \\
& \underbrace{|B_y \setminus \{i\}| \exp(\frac{1}{|B_y \setminus \{i\}|}\sum_{k\in |B_y \setminus \{i\}|} \langle z_i,z_k \rangle - \langle z_i,z_j \rangle)}_{confliction \; term}  \\
& + \underbrace{|B_y^C|\exp(\frac{1}{|B_y^C|}\sum_{k\in B_y^C} \langle z_i,z_k \rangle - \langle z_i,z_j \rangle)}_{confrontation \; term} )
\end{split}
\end{equation}
Such a lower bound for unsupervised contrastive learning (UCL) loss also follows from an application of Jensen’s inequality. To be specific, we bring the class-specific batch-wise loss in a form amenable to Jensen’s inequality:
\begin{equation}
\begin{split}
\mathcal{L}_{\text{uc}}(Z;Y,B,y) &=  \sum_{i\in B_y} 
\log (\frac{\exp(\langle z_i,z_j\rangle)}{\sum_{k \in B \setminus \{i\}} \exp(\langle z_i,z_k\rangle)) }) \\
& = \sum_{i\in B_y} \log( \frac{\sum_{k\in B \setminus \{i\}}\exp(\langle z_i,z_k\rangle)}{\exp(\langle z_i,z_j\rangle) } ) \\
&= \sum_{i\in B_y} \log( \frac{\sum_{k\in B \setminus \{i\}}\exp(\langle z_i,z_k\rangle)}{\exp(\langle z_i,z_j\rangle) } )
\end{split}
\label{equation:usc}
\end{equation}
Thus, we bound the argument of the log in Equation (\ref{equation:usc}) by:
\begin{equation}
\begin{split}
    &\mathcal{L}_{\text{uc}}(Z;Y,B,y) = \sum_{i\in B_y} \log( \frac{\sum_{k\in B \setminus \{i\}}\exp(\langle z_i,z_k\rangle)}{\exp( \langle z_i,z_j\rangle) } )\\
    & = \sum_{i\in B_y} \log( \frac{ \sum_{k\in B_y \setminus \{i\}}\exp(\langle z_i,z_k\rangle) + \sum_{k\in B_y^C}\exp(\langle z_i,z_k\rangle)}{\exp( \langle z_i,z_j\rangle) } ) \\
    & \ge \sum_{i\in B_y} \log( \\
& \underbrace{|B_y \setminus \{i\}| \exp(\frac{1}{|B_y \setminus \{i\}|}\sum_{k\in |B_y \setminus \{i\}|} \langle z_i,z_k \rangle - \langle z_i,z_j \rangle)}_{confliction \; term}  \\
& + \underbrace{|B_y^C|\exp(\frac{1}{|B_y^C|}\sum_{k\in B_y^C} \langle z_i,z_k \rangle - \langle z_i,z_j \rangle)}_{confrontation \; term} )
\end{split}
\end{equation}
Note that equality is attained if and only if the above conditions hold for every $i \in B_y$. Also, note that the respective constants, $C_i(B, y)$ and $D_i(B, y)$, depend indeed on the batch $B$ and the label $y$.

\begin{table*}
    \centering
    \setlength\tabcolsep{2.3pt}
    \begin{tabular}{l|cc|cccccc}
    \hline
        \multirow{3}{*}{Model Name}&\multicolumn{8}{c}{HAR} \\
        \cline{2-9}
        &\multicolumn{2}{c|}{Overall results} &\multicolumn{6}{c}{F1-score for each class}\\
        \cline{2-9}
        & Accuracy &F1-score & Walking(17\%) & Upstairs(15\%)&Downstairs(13\%) &Standing(17\%) &Sitting(19\%) &Lying(19\%)\\
        \hline
        Supervised  & 92.22±0.63 & 92.22±0.67 & \textbf{98.17±1.10}   & 95.77±1.64   & 94.72±2.38   & 80.11±1.68   & 84.56±1.10      & \textbf{100.00±0.00}\\
        SCL         & 86.30±2.65 & 86.26±2.68 & 90.07±2.72   & 85.23±7.25   & 87.79±1.57   & 76.56±2.69   & 80.65±2.37   & 97.27±1.77\\
        \hline
        SimCLR     &78.67±1.14 &78.01±0.94 & 79.09±1.94   & 70.70±2.00   & 71.24±0.50   & 67.56±0.77   & 79.30±0.92      & 80.19±1.35\\
        T-loss      &79.51±3.49 &78.78±3.83 & 71.83±5.71   & 72.85±5.75   & 72.43±8.92   & 76.81±3.45   & 79.64±2.30      & 99.11±1.07\\
        CPC         &68.56±3.04 & 68.32±3.22 & 47.31±8.26   & 49.01±5.94   & 74.59±7.67   & 78.94±1.84   & 61.51±6.64   & 98.56±1.64\\
        TSTCC       &88.31±1.00 &88.31±1.02 & 91.56±2.53   & 88.02±2.69   & 91.09±2.01   & 79.08±2.37   & 80.55±1.05   & 99.54±0.82\\
        TS2Vec      &\underline{92.98±0.49} &\underline{93.01±0.50} & 97.53±0.25   & \underline{95.98±1.01}   & \underline{95.49±0.93}   & \textbf{82.92±0.77}   & \textbf{86.25±0.69}      & \underline{99.85±0.17}\\
        \hline
        \textbf{SIP-LDL} &\textbf{93.20±0.66} &\textbf{93.27±0.64} & \underline{97.77±0.24}   & \textbf{96.94±0.21}   & \textbf{96.80±1.14}      & \underline{82.22±0.75}   & \underline{85.94±0.28}   & 99.66±0.24\\
        \hline
    \end{tabular}
\small\caption{The comparison of the state-of-the-art approaches on the HAR dataset. The best results are marked in bold, the second-place results are underlined. The proportion of samples in each class is represented in parentheses.}
\label{table:har}
\end{table*}

\begin{table*}
    \centering
    \setlength\tabcolsep{2.4pt}
    \begin{tabular}{l|cc|ccccc}
    \hline
        \multirow{3}{*}{Model Name}&\multicolumn{7}{c}{Sleep-EDF} \\
        \cline{2-8}
        &\multicolumn{2}{c|}{Overall results} &\multicolumn{5}{c}{F1-score for each class}\\
        \cline{2-8}
        & Accuracy &F1-score &Wake (19\%)&N1 (5\%)&N2 (42\%)&N3 (15\%)&REM (19\%)\\
        \hline
        Supervised  & 83.29±0.62 &74.26±0.43 & 91.74±0.80   & \textbf{29.83±1.86}   & 88.75±0.39  & 89.97±1.11   & \underline{77.30±0.52}\\
        SCL         &\underline{84.23±0.38} &\textbf{75.01±0.30} & \textbf{92.94±0.37}    &23.55±1.85   & 88.67±0.36   & 90.03±0.78   & \textbf{78.15±0.34}\\
        \hline
        SimCLR     &64.63±1.41 &54.21±1.15 & 79.39±9.25   & 15.50±0.48    & 76.60±2.58   & 86.50±1.24   & 55.54±0.76\\
        T-loss      &80.53±0.78 &70.15±0.32  &  84.49±1.86   & 17.29±1.77   & 87.96±0.97   &  88.89±1.31   &  73.13±0.72 \\
        CPC         &62.68±2.62 &47.71±3.10  & 74.78±2.46   & 10.67±0.88    & 68.54±2.57   & 75.70±9.71   & 14.87±6.26\\
        TSTCC       &81.41±0.55 &70.53±0.29 & 86.59±0.89   & 13.58±0.76   & \textbf{89.65±0.84}   & \textbf{91.87±0.33}   & 71.74±0.82 \\
        TS2Vec      &80.46±0.41 & 69.04±0.55 & 89.03±0.16   & 10.10±1.85   & 84.80±0.51   & 86.20±0.53   & 71.85±0.94\\
        \hline
        \textbf{SIP-LDL} &\textbf{84.32±0.35} & \underline{74.46±0.41} & \underline{92.60±0.41}   & \underline{25.74±1.45}      & \underline{89.11±0.48}   &\underline{91.89±1.44}   & 77.05±1.39\\
        \hline
    \end{tabular}
\small\caption{The comparison of the state-of-the-art approaches on the Sleep-EDF dataset. The best results are marked in bold, the second-place results are underlined. The proportion of samples in each class is represented in parentheses. }
\label{table:sleef-edf}
\end{table*}

\begin{table*}
    \centering
    \setlength\tabcolsep{2.4pt}
    \begin{tabular}{l|cc|cccc}
    \hline
        \multirow{3}{*}{Model Name}&\multicolumn{6}{c}{PhysioNet 2017} \\
        \cline{2-7}
        &\multicolumn{2}{c|}{Overall results} &\multicolumn{4}{c}{F1-score for each class}\\
        \cline{2-7}
        & Accuracy &F1-score &AF (9\%)&Normal (59\%)&Other (29\%)&Noisy (3\%)\\
        \hline
        Supervised  & 54.00±3.22 &38.16±1.11  & 16.62±3.81   & 66.88±10.60  &  \textbf{41.94±4.40}   & \underline{34.15±1.99}\\
        SCL         & 56.42±0.11 & 21.96±0.63 & 2.33±0.16    & 69.02±0.07   & 14.16±1.39    & 14.83±2.11\\
        \hline
        SimCLR     &56.29±0.08 & 21.75±0.88 & 1.68±0.60    & 68.06±0.06   &4.20±1.10    & 8.52±3.72\\
        T-loss     &54.27±0.84 & 37.15±0.51 &12.58±2.04   & 67.20±1.35   & 36.85±3.84   & 29.29±1.08\\
        CPC        &54.46±1.27 & 38.83±0.52 & 19.66±1.81   & 66.49±1.66   & 35.79±3.39   & 30.91±1.41\\
        TSTCC      &\underline{56.52±1.02}& \underline{40.48±0.50}  &\underline{20.80±0.30}   & \underline{69.52±1.26}   & 36.15±1.54   & 31.23±1.59\\
        TS2Vec     &56.27±0.24 & 24.39±1.11 & 1.33±0.35    & 71.75±0.12   &  6.76±3.80   & 12.82±1.92 \\
        \hline
        \textbf{SIP-LDL} &\textbf{58.71±0.54} & \textbf{41.32±1.24} & \textbf{21.21±1.82}   & \textbf{72.80±1.02}   & \underline{39.63±0.95}    & \textbf{34.22±3.00}  \\
        \hline
    \end{tabular}
\small\caption{The comparison of the state-of-the-art approaches on the PhysioNet 2017 dataset. The best results are marked in bold, the second-place results are underlined. The proportion of samples in each class is represented in parentheses.}
\label{table:pyhsionet2017}
\end{table*}

\begin{table*}
    \centering
    \setlength\tabcolsep{2.4pt}
    \begin{tabular}{l|cc|cccc}
    \hline
        \multirow{3}{*}{Model Name}&\multicolumn{6}{c}{TUSZ} \\
        \cline{2-7}
        &\multicolumn{2}{c|}{Overall results} &\multicolumn{4}{c}{F1-score for each class}\\
        \cline{2-7}
        & Accuracy &F1-score &CF (80\%)&GN (16\%)&AB (3\%)&CT (1\%)\\
        \hline
        Supervised  &63.42±0.64 &41.67±2.11  & \textbf{79.51±1.58}   & \textbf{10.78±3.25}  & 74.07±3.99   & 3.23±1.94\\
        SCL         & 58.81±0.93 & 26.62±4.26  & 73.97±0.54   & 3.00±0.04    & 25.24±15.58  & 1.52±3.65\\
        \hline
        SimCLR     &56.89±0.09 &18.29±0.32    & 72.46±0.06   & 2.01±0.03    & 0.00±0.00    & 0.63±1.27\\
        T-loss      &62.03±0.53 &38.07±0.96   & 77.29±0.50   & 3.97±3.24    & 72.73±2.15   & 0.00±0.00\\
        CPC         &61.77±0.43 &36.41±0.71   & 75.42±0.21   &  0.33±0.66    &  63.89±3.57   &  6.39±2.54\\
        TSTCC       &60.73±1.03 &35.36±3.98   & 74.87±0.57   & 1.87±1.80    & 53.73±10.26  & 3.28±2.59\\
        TS2Vec      &62.50±0.77 & 42.26±1.61  & 77.64±0.44   & 5.59±1.43    &  \underline{77.50±1.22}   & \textbf{19.44±4.98}\\
        \hline
        \textbf{SIP-LDL} &\textbf{63.72±0.99} &\textbf{48.20±3.23} & \underline{79.22±0.77}   & \underline{9.75±2.82}    & \textbf{79.55±12.91}     & \underline{19.18±7.67}\\
        \hline
    \end{tabular}
\small\caption{The comparison of the state-of-the-art approaches on the TUSZ dataset. The best results are marked in bold, the second-place results are underlined. The proportion of samples in each class is represented in parentheses.}
\label{table:tusz}
\end{table*}

\section{F. Additional Evaluation Results}
We compare our SIP-LDL framework with the other prevailing baseline methods for human activity recognition, sleep stage classification, cardiac arrhythmia classification, and epileptic seizure classification on Table \ref{table:har}-\ref{table:tusz}. Specifically, we present the overall F1 and the F1 for each class as the additional evaluation results of the overall performance in the main manuscript.

We can find that the current contrastive learning methods exhibit significant differences in performance across different classes. Our proposed method, SIP-LDL, can serve as an extension of SimCLR, significantly improving performance, especially for the minority classes, without introducing additional parameters.
Specifically, we outperform current UCL methods on almost all minority classes on three imbalanced datasets (e.g., N1 in Sleep-EDF, AF in PhysioNet 2017, AB in TUSZ) while approximating the supervised methods on specific minority classes.